%% file: ijcai22.tex
\pdfoutput=1
\documentclass{article}
\usepackage{ijcai22}

\usepackage{soul}
\usepackage{url}
\usepackage[hidelinks]{hyperref}
\usepackage[utf8]{inputenc}
\usepackage[small]{caption}
\usepackage{graphicx}
\usepackage{amsmath}
\usepackage{amsthm}
\usepackage{booktabs}

\usepackage{booktabs}
\usepackage{url}

\usepackage{multirow}

\usepackage{amsfonts}
\usepackage{thmtools}

\theoremstyle{theorem}
\newtheorem{theorem}{Theorem}

\newtheorem{definition}[theorem]{Definition}

\AtBeginDocument{\providecommand\BibTeX{{\normalfont B\kern-0.5em{\scshape i\kern-0.25em b}\kern-0.8em\TeX}}}

\usepackage{arydshln}

\usepackage{color, colortbl}
\definecolor{Gray}{gray}{0.9}

\newcolumntype{g}{>{\columncolor{Gray}}c}

\usepackage{thm-restate}
\usepackage[mathic=true]{mathtools}
\usepackage{fixmath}

\usepackage{pifont}

\usepackage{enumitem}
\setlist[enumerate]{noitemsep, topsep=0.5\topsep}
\setlist[description]{noitemsep, topsep=0.5\topsep}
\setlist[itemize]{noitemsep, topsep=0.5\topsep}

\usepackage{setspace}
\usepackage[protrusion=true,expansion=false, activate={true,nocompatibility},final,kerning=true,spacing=true,stretch=0, shrink=80]{microtype}

\usepackage{times}

\newcommand{\citet}[1]{\citeauthor{#1}~\shortcite{#1}}

\title{A Unified View of Relational Deep Learning for Drug Pair Scoring}

\author{
Benedek Rozemberczki$^1$\and Stephen Bonner$^2$ \and Andriy Nikolov$^1$ \and Michaël Ughetto$^1$ \and\\ Sebastian Nilsson$^1$ \And Eliseo Papa$^1$
\affiliations
$^1$Research Data \& Analytics, Research \& Development IT, AstraZeneca\\
$^2$Data Sciences and Quantitative Biology, Discovery Sciences R\&D, AstraZeneca\\
\emails
benedek.rozemberczki@astrazeneca.com
}

\begin{document}

\maketitle

\begin{abstract}
In recent years, numerous machine learning models which attempt to solve polypharmacy side effect identification, drug-drug interaction prediction and combination therapy design tasks have been proposed. Here, we present a unified theoretical view of relational machine learning models which can address these tasks. We provide fundamental definitions, compare existing model architectures and discuss performance metrics, datasets and evaluation protocols. In addition, we emphasize possible high impact applications and important future research directions in this domain.
\end{abstract}

\input{sections/introduction}

\input{sections/background}

\input{sections/models}

\input{sections/evaluation}

\input{sections/applications}

\input{sections/discussion_and_future_directions}

\input{sections/conclusion}

\small

\section*{Acknowledgement}
The authors would like to thank Peizhen Bai, Piotr Grabowski, Haiping Lu, Rocío Mercado, and Paul Scherer for help and feedback throughout the preparation of this manuscript. Stephen Bonner is a fellow of the AstraZeneca postdoctoral program.

\bibliographystyle{named}
\bibliography{bibliography}

\end{document}

%% file: sections/introduction.tex
\section{Introduction}
Relational deep learning has an unprecedented potential for revolutionizing the drug discovery process and pharmaceutical industry \cite{gaudelet2021utilizing}. A number of high value use cases for relational deep learning in the pharmaceutical domain involve answering questions about what happens when two drugs are administered at the same time. These potential applications might want to answer questions such as: Will a combination of two drugs be more effective at destroying a specific type of lung cancer cells \cite{preuer2018deepsynergy}? Is there an unexpected (polypharmacy) side effect \cite{zitnik2018modeling} of using these two drugs together? Is there an unwanted chemical interaction \cite{kwon2017deepcci} that these drug molecules can have?

All of these previously mentioned questions can be answered by what we see as \textit{drug pair scoring}, a machine learning task that involves a set of drugs and the task of predicting the behaviour of pairs in a specific context of interest. Given an incomplete database of drug pairs, drug administration contexts and outcomes, the goal is to train a model to accurately make probabilistic predictions for unseen entries. The reasons for answering these questions via algorithmic methods are multi-fold. Firstly, testing all drug pairs in all of the contexts is not feasible due to time and financial constraints such as drug prices and labour costs \cite{preuer2018deepsynergy}. Secondly, certain pair scoring tasks such as polypharmacy side effect prediction can only be validated in human-based trials. Finally, laboratory testing of drug pairs is prone to human errors \cite{liu2020drugcombdb}.

Traditional supervised machine learning methods which solve the drug pair scoring task use handcrafted molecular features to predict the outcome of administering the drugs together in a specific context \cite{sidorov2019predicting,chiang2020drug}. Another group of techniques uses an unsupervised approach which diffuses the profile of the drug pair on a heterogeneous biological graph
\cite{zhang2017predicting,li2018network,huang2019driver} in order to find potential polypharmacy, synergy or interaction indications. Deep learning techniques which solve the drug pair scoring task can be seen as a fusion and extension of these traditional methods. Such models first generate drug representations based either on molecular structure or the heterogeneous graph based neighbourhood context. In the second optional step, these representations are propagated in the biological graph and aggregated. Finally, drug pair representations are formed and probability scores are outputted in the specific drug administration contexts. We present a high level summary of the drug pair scoring task idea in Figure \ref{fig:candy}.

\input{./figures/eyecandy.tex}

Our main contributions can be summarized as:
\begin{enumerate}
    \item We provide a unification of drug-drug interaction, polypharmacy side effect and synergistic drug combination prediction tasks. 
    \item We present an overview on the design of relational machine learning models which can address these predictive tasks. 
    \item We highlight the publicly available datasets used to train and test the models on these tasks and survey the literature for the most commonly used evaluation metrics.
    \item We review the most important applications of these techniques and discuss directions for future research in the domain.
\end{enumerate}
The remainder of this survey is structured as follows. In Section \ref{sec:background} we establish the foundations of a unified view of discriminative machine learning tasks defined on pairs of drugs. Section \ref{sec:models} discusses the architectural details of models that can solves these tasks. The evaluation metrics, protocols and datasets used in the literature are detailed in Section \ref{sec:evaluation}. Several important key application areas are highlighted in Section \ref{sec:applications}. We discuss the limitations of current approaches and future research directions in Section \ref{sec:discussion}. The paper concludes with Section \ref{sec:conclusions}. The survey is supported by a collection of relevant works under the \url{https://github.com/AstraZeneca/polypharmacy-ddi-synergy-survey} repository.

%% file: figures/eyecandy.tex
\begin{figure}[h!]
\centering 
\includegraphics[scale=0.15]{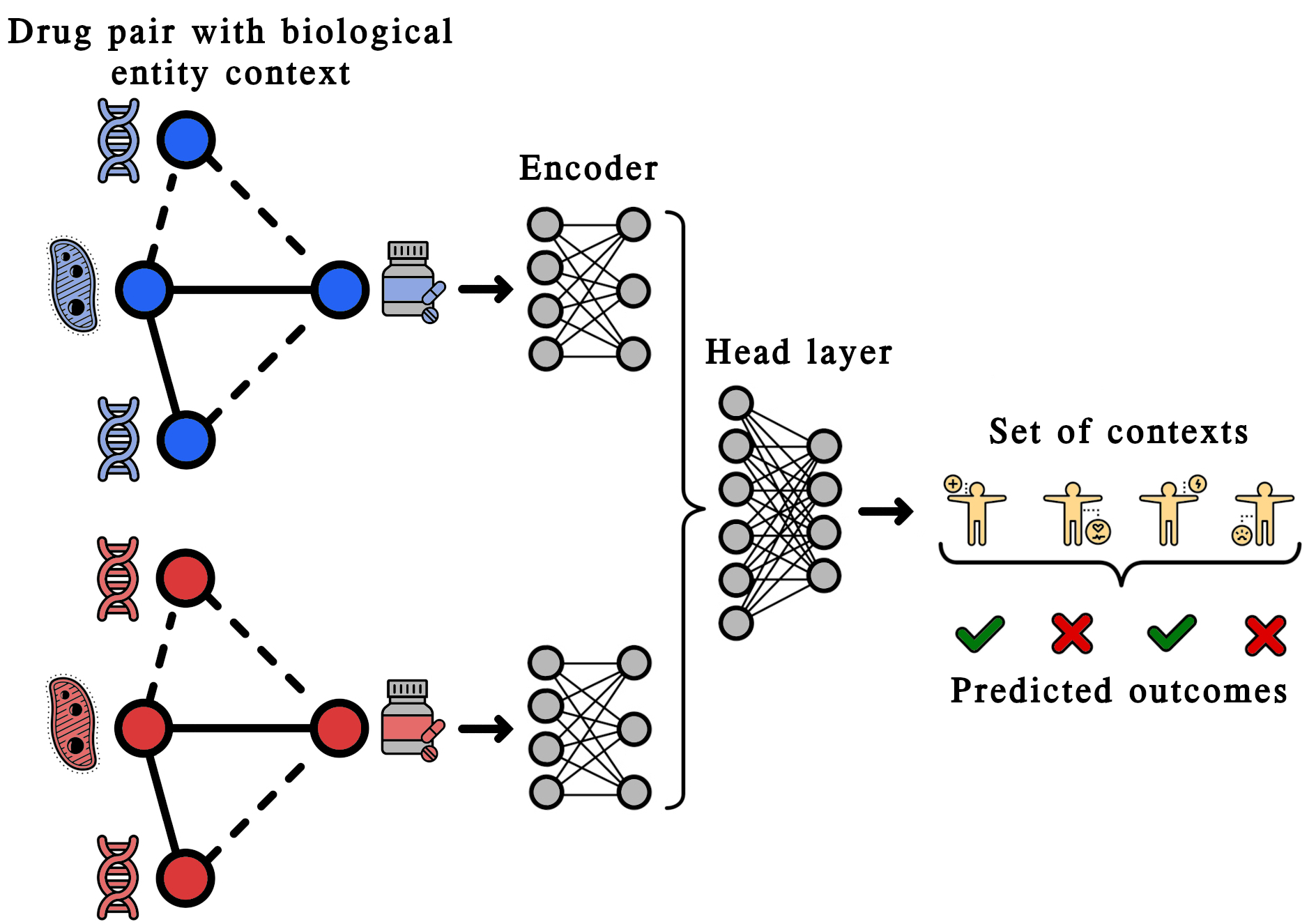}
\caption{Drug-drug interaction, polypharmacy side effect, and pair combination therapy design prediction tasks follow the same template. Given a pair of drugs with optional biological context, the task is to predict an outcome in a specific application domain. Relational machine learning models which solve these task can exploit molecular features, knowledge graph based neighbourhoods or both. }\label{fig:candy}
\end{figure}

%% file: sections/background.tex
\section{Background}\label{sec:background}
Our discussion of drug pair scoring models requires the introduction of a drug set $\mathcal{D}=\{d_1,\dots,d_n \}$ that describes compounds of interest and a context set $\mathcal{C}=\{c_1,\dots, c_k\}$ that contains contexts where two drugs are used in a pair combination.
\begin{definition}
\textbf{Labeled drug pair.} A labeled drug pair defined on drug set $\mathcal{D}$ and context set $\mathcal{C}$ is the tuple $(d,d',c,y_{d,d',c})$, where the binary indicator $y_{d,d',c}\in\left\{0,1\right\}$ is the outcome for drug pair $d,d'\in \mathcal{D}$ in context $c\in\mathcal{C}$.
\end{definition}
A labeled drug pair is a known fact about the drug pair having an effect in a context such as a specific polypharmacy side effect, interaction or synergistic relationship at treating a disease. The purpose of pair scoring models is to learn from these tuples to predict the labels for unlabeled drug pairs and contexts.
\begin{definition}
\textbf{Database of labeled drug pairs.} A database of labeled drug pairs defined on drug and context sets $\mathcal{D}$ and $\mathcal{C}$ is the set $\mathcal{S}$ containing labeled drug pairs $(d,d',c,y_{d,d',c})$ where  $d,d'\in \mathcal{D}$ and $c\in\mathcal{C}$.
\end{definition}
Pair scoring models are trained on databases of labeled drug pairs and the trained models are used to predict the label of pairs for which we do not know the outcome in certain contexts.

\begin{definition}
\textbf{Heterogeneous interaction graph with drug entities.} 
 We denote with $G(\mathcal{V},\mathcal{R},\mathcal{E})$ the heterogeneous interaction graph with drug entities,  where $\mathcal{V}$ and $\mathcal{R}$ are the entity and relation sets, it holds that the drug set $\mathcal{D} \subset \mathcal{V}$ and $\mathcal{E}$ is formed by typed edges of the form  $(v,r, u)\in \mathcal{V} \times \mathcal{R} \times \mathcal{V}$ . 
\end{definition}
We consider a heterogeneous graph where the drug set is a subset of the vertex set. This definition of heterogeneous (biological) knowledge graph helps to create knowledge graph based representation for the compounds of interest.
\begin{definition}\label{def:neigh}
\textbf{Neighbourhood encoder.} A neighbourhood encoder is the  function: 
\begin{align}
\mathbf{h}_d=\textsc{aggregate}(\Theta_u, \forall u\in \mathcal{N}(d))\label{eq:encoder}.
\end{align}
In Equation \eqref{eq:encoder}  $\Theta_u$ is a parametric vector representation of $ u \in \mathcal{V}$ and $\mathcal{N}(\cdot)$ is a neighbourhood set.
\end{definition}
The neighbourhood encoder function \cite{hamilton2017inductive} creates a vector representation of drug vertices of the graph  based on the aggregation of trainable parameter vectors in the neighbourhood of the source node. Neighbourhoods of a drug can be defined based on arbitrary notions of proximity and the aggregation itself could be a parametric transformation. 

\begin{definition}\label{def:mol}
\textbf{Molecular encoder.} A molecular encoder is the function $\textbf{h}_d=h_{\Theta}(\mathcal{M}_d)$, parametrized by $\Theta$ where $\textbf{h}_d$ is the learned vector representation and $\mathcal{M}_d$ is a generic notation of molecular features describing the drug $d$.
\end{definition}
A molecular encoder is a neural network which generates a vector representation from the features of the molecule - these molecular features can be derived from generic features (e.g. hydrophilicity), a string representation, molecular graph or geometry. 
\begin{definition}
\textbf{Neighbourhood informed molecular encoder.}\label{def:informed}
This encoder is the function : $$\textbf{h}_d=\textsc{aggregate}(h_{\Theta}(\mathcal{M}_u), \forall u\in \mathcal{N}(d));$$ where $h_{\Theta}(\mathcal{M}_u)$ and $\textsc{aggregate}(\cdot, \forall u\in \mathcal{N}(d))$ are molecular and neighbourhood encoders respectively.
\end{definition}
This encoder combines the layers described in Definitions \ref{def:neigh} and \ref{def:mol}. It is essentially a neighbourhood encoder parametrized by representations outputted by a molecular encoder -- molecular representations learned by the molecular encoder are aggregated in the neighbourhood of source drug nodes in the knowledge graph which has drug entities.
\begin{definition}
\textbf{Molecular representation combiner.} \label{def:combiner} Given the drugs $d,d^{\prime} \in \mathcal{D}$ with vector representations $\textbf{h}_,\textbf{h}_{d^{\prime}}$ the molecular representation combiner is the function $\textbf{h}_{d,d'}=g _{\Theta}(\textbf{h}_d;\textbf{h}_{d'})$ that outputs $\textbf{h}_{d,d'}$ a vector representation of the drug pair.
\end{definition}
The representation output by this combiner function can be drug orchestration order dependent. This way the temporal order of drug orchestration can be expressed by the pair scoring model. For example the concatenation of drug vectors results in order dependent representations of pairs, while a bilinear transformation of drug representations with a diagonal matrix does not.
\begin{definition}
\textbf{Scoring head layer.} \label{def:head}The scoring head layer is the function $\hat{y}_{d,d',c}=k_{\Theta}(\textbf{h}_{d,d'},c)$, where  the predicted label proability satisfies that $\{\hat{y}_{d,d',c} \in \mathbb{R}\mid 0\leq \hat{y}_{d,d',c}\leq 1\}$.
\end{definition}
Given a drug pair representation and a context, the scoring head layer outputs a probability score for the outcome.
\begin{definition}
\textbf{Drug pair scoring loss and cost functions.} \label{def:cost}Given the drug pair $d,d'\in \mathcal{D}$ , context $c\in \mathcal{C}$, ground-truth label $y_{d,d',c}$  and predicted score $\hat{y}_{d,d',c}$ the loss is defined as the function $\ell(y_{d,d',c};\hat{y}_{d,d',c})$. The cost on the whole drug pair database $\mathcal{S}$ is defined by Equation \eqref{eq:losses}.
\begin{align}
\mathcal{L} &= \sum \limits_{(d,d',c) \in \mathcal{S}}\ell(y_{d,d',c};\hat{y}_{d,d',c}).\label{eq:losses}
\end{align}
\end{definition}
In practical settings, drug pair scoring models are trained by the minimization of the binary cross-entropy summed over the labeled drug pair, context triples.

%% file: sections/models.tex
\input{./tables/models.tex}

\section{Drug Pair Scoring Models}\label{sec:models}
Our discussion of the drug pair scoring models introduces our unified view about the general architecture of these models and compares state-of-the-art architecture designs. 
\subsection{Unified View: The Drug Pair Scoring Model}

Based on the definitions outlined in Section \ref{sec:background} we propose a unified view of drug pair scoring models. We postulate that the abstract design of drug pair scoring models irrespective of the specific subtask solved always has the following  architecture:
\begin{enumerate}
    \item An encoder to generate drug representations - this can be one of the functions described by Definitions \ref{def:neigh},  \ref{def:mol} and \ref{def:informed}.
    \item A molecular representation combiner function to generate a drug pair representation -- see Definition \ref{def:combiner}.
    \item The scoring head layer to predict the probability of a context dependent outcome proposed by Definition \ref{def:head}.
    \item The loss function of Definition \ref{def:cost} which depends on ground-truth labels and the probabilities output by the head layer. 
\end{enumerate}
This architecture and design allows for the joint end-to-end training of the individual model components -- gradient descent based update of the layer weights.
\subsection{Specific Architecture Designs}
We compare state-of-the-art model architectures in Table \ref{table:general} that can solve pair scoring tasks. Our comparison considers the model level, induction capabilities, specific subtask, node types of the heterogeneous graph and the molecular features exploited by the model. Model attributes used for comparison were the following:

\begin{itemize}
    \item \textit{Model level:} A model operates at the following levels based on the encoder architecture used for generating the drug representations: (a) higher-view -- \textit{neighbourhood encoder}, (b) lower-view -- \textit{molecular encoder}, (c) hierarchical-view -- \textit{neighbourhood informed molecular encoder}.
    \item \textit{Machine learning task:} The drug pair scoring task of interest solved by the dedicated model architecture proposed in the research paper. It has to be one of interaction, polypharmacy or synergy prediction.
    \item \textit{Induction:} A model is inductive if it can predict the label of drug pairs where at least one of the drugs was not in the training set drug pairs.
    \item \textit{Entities:} The types of hetereogeneous graph entities (\textit{drugs}, \textit{proteins}, \textit{diseases}) used by the model to solve the task.
    \item \textit{Drug features:} Molecular features and information about the compound encoded by the molecular encoder function.
\end{itemize}
Our comparison highlights that there is a hard trade-off between induction and the exclusion of compound features. It is also evident that there is a connection between the machine learning subtask and the model architecture design: for example, polypharmacy side effect prediction models are mostly high level transductive neighbourhood encoders with a scoring layer on top. Synergy scoring models are mostly inductive techniques which exploit the molecular information about the drugs. Currently, there is no single pair scoring model which includes all of the considered biological modalities.






%% file: tables/models.tex
\begin{table*}[h!]

\captionsetup{width=1.0\linewidth}
\centering
\caption{A machine learning task, model view level, induction, interaction graph node type (entity) and drug feature based comparison of drug pair scoring machine learning models. Machine learning models that solve a specific pair scoring task are ordered chronologically in the table.}\label{table:general}

{\tiny
\begin{tabular}{c g c  g c ggg cccc}

 \multicolumn{1}{c}{}& \multicolumn{1}{c}{} & \multicolumn{1}{c}{}  &\multicolumn{1}{c}{}&\multicolumn{1}{c}{} &\multicolumn{3}{c}{\textbf{Entity Types}} & \multicolumn{4}{c}{\textbf{Drug Features}}\\ \hline
 \textbf{Task}&\textbf{Method}&\textbf{Reference}&\textbf{Model view}& \textbf{Induction}&
\textbf{Drug}&\textbf{Protein}&\textbf{Disease}&
\textbf{SMILES}&\textbf{Graph}&
\textbf{Geometry}&\textbf{Generic}\\
\hline
\multirow{6}{*}{\textbf{Polypharmacy}}&DECAGON   &\cite{zitnik2018modeling}        &Higher &&$\bullet$&$\bullet$&&&&&\\
&KBLRN  &\cite{malone2018knowledge}        &Higher  &&$\bullet$&$\bullet$&&&&&\\
 &SDHINE     &\cite{hu2018adverse}        &Higher  &&$\bullet$&$\bullet$&$\bullet$&&&&\\
&ESP       &\cite{burkhardt2019esp}        &Higher  &&$\bullet$&$\bullet$&&&&&\\
&MHCADDI   &\cite{deac2019mhcaddi}& Lower  &$\bullet$&&&&&$\bullet$&&\\ 
&TIP       &\cite{xu2020tip}                &Higher  &&$\bullet$&$\bullet$&&&&&\\
\hline
\multirow{22}{*}{\textbf{Interaction}}&DeepCCI   &\cite{kwon2017deepcci}          & Lower  &$\bullet$&&&&$\bullet$&&&$\bullet$\\
&MVGAE &    \cite{ma2018drug}       & Hierarchical&$\bullet$&$\bullet$&$\bullet$&$\bullet$&$\bullet$&&&\\
&DeepDDI   &\cite{ryu2018deep}            & Lower  &$\bullet$&&&&$\bullet$&&&\\
&D$^3$I      &\cite{peng2019d3i}        &Hierarchical  &&$\bullet$&&&&&&$\bullet$\\
&MR-GNN   &\cite{xu2019mr}& Lower  &$\bullet$&&&&&$\bullet$&&\\
&SkipGNN   &\cite{huang2020skipgnn}        &Higher  &&$\bullet$&$\bullet$&$\bullet$&&&&\\
&CASTER&\cite{huang2019caster}            & Lower  &$\bullet$&&&&$\bullet$&&&\\
&DeepDrug&\cite{cao2020deepdrug}            & Lower  &$\bullet$&&&&$\bullet$&$\bullet$&$\bullet$&\\
&GoGNN&\cite{gognn}            &Hierarchical  &&$\bullet$&&&&$\bullet$&&\\
&DPDDI   &\cite{feng2020dpddi}          & Lower  &$\bullet$&&&&&&&$\bullet$\\
&KGNN &\cite{kgnn}            &Higher &&$\bullet$&$\bullet$&$\bullet$&&&&\\
&BiGNN &\cite{bai2020bi}            &Hierarchical &$\bullet$&$\bullet$&$\bullet$&&&$\bullet$&&\\
&MIRACLE&\cite{wang2021multi}            &Hierarchical &$\bullet$&$\bullet$&&&&$\bullet$&&\\
&EPGCN-DS &\cite{sun2020structure}      & Lower&$\bullet$&&&&&&&\\
&SumGNN &\cite{yu2021sumgnn}            &Higher &&$\bullet$&$\bullet$&$\bullet$&&&&\\
&DANN-DDI&\cite{dannddi}            &Higher &&$\bullet$&$\bullet$&&&&&\\
&SSI-DDI  &\cite{nyamabossi}& Lower  &&&&&&$\bullet$&&\\ 
&MTDDI  &\cite{mtddi2021}&Hierarchical  &&$\bullet$&$\bullet$&$\bullet$&&&&$\bullet$\\ 
&MUFFIN &\cite{chen2021muffin}&Hierarchical  &&$\bullet$&$\bullet$&$\bullet$&$\bullet$&&&\\ 
&DDIAAE &\cite{dai2021drug}&Higher &&$\bullet$&$\bullet$&$\bullet$&&&&\\ 
&RWGCN &\cite{feeney2021relation}&Higher &&$\bullet$&$\bullet$&$\bullet$&&&&\\ 
&SmileGNN &\cite{han2021smilegnn}&Hierarchical &&$\bullet$&$\bullet$&&$\bullet$&&&\\ 
&GCN-BMP &\cite{CHEN202047gcnbmp}& Lower &$\bullet$&$\bullet$&&&$\bullet$&&&\\ 
\hline 
\multirow{11}{*}{\textbf{Synergy}}&DeepSynergy   &\cite{preuer2018deepsynergy}        & Lower  &$\bullet$&&&&&&&$\bullet$\\
&MCDC &\cite{mcdc}        & Hierarchical  &&$\bullet$&$\bullet$&&$\bullet$&&&\\
&DTF  &\cite{sun2020dtf}        & Higher  &&$\bullet$&&&&&&\\
&DeepSignalFlow &\cite{zhang2021mining}& Hierarchical&&$\bullet$&$\bullet$&$\bullet$&&&&\\ 
&DeepDDS&\cite{wang2021deepdds}& Lower&$\bullet$&&&&$\bullet$&&&\\ 
&GraphSynergy&\cite{yang2021graphsynergy}& Higher&&$\bullet$&$\bullet$&&&&&\\ 
&TranSynergy&\cite{liu2021transynergy}& Hierarchical&$\bullet$&$\bullet$&$\bullet$&&&&&$\bullet$\\ 
&MatchMaker&\cite{brahim2021matchmaker}& Lower&$\bullet$&&&&$\bullet$&&&$\bullet$\\ 
&AuDNNSynergy&\cite{zhang2021synergistic}& Lower&$\bullet$&&&&$\bullet$&&&\\ 
&AID&\cite{kim2021anticancer}& Hierarchical&$\bullet$&$\bullet$&$\bullet$&&$\bullet$&&&\\
&MOOMIN &\cite{rozemberczki2021moomin}&Hierarchical  &$\bullet$&$\bullet$&$\bullet$&&&$\bullet$&&\\
\bottomrule
\end{tabular}
}

\end{table*}

%% file: sections/evaluation.tex
\section{Evaluation}\label{sec:evaluation}
The evaluation of machine learning models requires performance metrics, train-test split strategies and publicly accessible datasets.
\subsection{Performance Metrics}
The predictive performance of drug pair scoring models is evaluated by metrics tailored to binary classification tasks. We summarise how these metrics are used for the evaluation of state-of-the-art drug pair scoring architectures in Table \ref{table:metrics}. 

\input{./tables/new_metrics.tex}

Looking at Table \ref{table:metrics} it is evident that the evaluation metrics used in the literature can be grouped into two categories:

\begin{itemize}
    \item \textit{Score based metrics:} These quantify predictive performance based over the whole domain of discrimination thresholds. The precision-recall area under the curve (AUPRC) considers the precision-recall trade off under the whole domain of discrimination thresholds while the receiver operating characteristic area under the curve (AUROC) considers false and true positive rates.
    \item \textit{Hard cut off evaluation metrics:} These performance metrics (accuracy, $\text{F}_1$ score, precision, recall) apply a hard discrimination threshold to assign a label to the data points based on the scores output by the pair scoring model. In order to calculate these, one needs to set a discrimination threshold.
\end{itemize}

Our findings demonstrate that pair scoring models are predominantly evaluated by score based metrics (AUPRC and AUROC) which do not require manual setting of a discrimination threshold. It is also evident that seminal research works which defined the key pair scoring tasks influenced the later evaluation metric choices -- polypharmacy prediction models adapted the  evaluation metrics from \cite{zitnik2018modeling} for example.
\subsection{Train-Test Split Strategies}
The evaluation of drug pair scoring tasks allows for the use of various train-test split strategies \cite{preuer2018deepsynergy} to test the performance of the model under cold-start and inductive scenarios \cite{dewulf2021cold}. Given a labeled drug pair-context database $\mathcal{S}$, defined on the drug and context sets $\mathcal{D}$ and $\mathcal{C}$, we assume that one can create the randomized splits $\mathcal{S}_{Train}$ and $\mathcal{S}_{Test}$. We summarized these splitting strategies in Figure \ref{fig:splits}.

\input{./figures/split.tex}

Using the formalism established to describe the pair scoring models, the splitting strategies are defined as: 
\begin{itemize}
\item \textbf{Random split:} labeled drug pair - context entries of $\mathcal{S}$ are randomly split between $\mathcal{S}_{Train}$ and $\mathcal{S}_{Test}$. 
\item \textbf{Drug pair stratified split:} A drug pair $d, d'\in \mathcal{D}$ that appears in entries of $\textit{S}_{Train}$ does not appear in entries of $\mathcal{S}_{Test}$. This split requires a pair scoring model which is inductive with respect to drugs.
\item \textbf{Drug stratified split:}  A drug $d \in \mathcal{D}$ that appears in entries of $\mathcal{S}_{Train}$ does not appear in entries of $\mathcal{S}_{Test}$. Like the drug pair stratified split this requires the model to be inductive with respect to new drugs.
\item \textbf{Context stratified split:} A context $c \in \mathcal{C}$ that appears in entries of $\mathcal{S}_{Train}$ does not appear in entries of $\mathcal{S}_{Test}$. This requires that the pair scoring model is inductive with respect to the set of contexts.
\end{itemize}
\subsection{Datasets}

We detail public sources for drug pair data which have been used by the approaches in this review in Table \ref{table:datasets}. Datasets are listed chronologically according to subtask and the licence and any restrictions for commercial use are detailed where available. It can be seen that the majority of datasets contain a small number of drugs, indicating most focus on approved drugs rather than all possible compounds, with the interactions captured in drug pairs being much more numerous.

\input{tables/datasets}

It should be noted that established resources such as TWOSIDES and DrugBank are frequently filtered, cleaned and split into new datasets. For example the Therapeutics Data Commons (TDC) resource contains filtered versions of both of these datasets designed for benchmark use~\cite{huang2021therapeutics}. It is also common for datasets to be named differently in publications, for example the split of TWOSIDES contained in TDC is also called ChChSe-Decagon in some works~\cite{biosnapnets}.

%% file: tables/new_metrics.tex
\begin{table}[h!]
\centering

\caption{Predictive performance evaluation metrics used by the research papers which proposed novel drug pair scoring techniques. Models are grouped by the pair scoring task and ordered chronologically.}\label{table:metrics}
{\tiny
\setlength{\tabcolsep}{1.5pt}
\begin{tabular}{lccccccc}
&
&  \multicolumn{6}{c}{\textbf{Evaluation metric}}\\
\cline{3-8}
\textbf{Model} &\textbf{Reference}& \textbf{AUPRC}& \textbf{AUROC} & \textbf{Precision} & \textbf{Recall}& \textbf{Accuracy} & $\textbf{F}_1$ \\\hline
     \textbf{DECAGON} &       \cite{zitnik2018modeling}  &          $\bullet$     &              $\bullet$    &        $\bullet$    &    &&       \\
     \textbf{KBLRN} &       \cite{malone2018knowledge}  &        $\bullet$     &              $\bullet$    &        $\bullet$    &   &&      \\
 \textbf{SDHINE}    &       \cite{hu2018adverse} &     &               $\bullet$    &        $\bullet$    &     &&    \\
      \textbf{ESP}    &       \cite{burkhardt2019esp} &      $\bullet$     &              $\bullet$    &        $\bullet$    &     &&   \\
 \textbf{MHCADDI}     &    \cite{deac2019mhcaddi}     &   &       $\bullet$      &                      &      &&   \\
 \textbf{TIP}   & \cite{xu2020tip}       &          $\bullet$     &              $\bullet$    &        $\bullet$    &   &&    \\
 \hline
 \textbf{DeepCCI}    &   \cite{kwon2017deepcci}      &        &              $\bullet$   &        &        &$\bullet$ &$\bullet$    \\
      \textbf{MVGAE}    &       \cite{ma2018drug}  &     $\bullet$    &        $\bullet$        &       &     &&  \\
     \textbf{DeepDDI}    &       \cite{ryu2018deep}  &       &                      &      $\bullet$   &  $\bullet$&$\bullet$ &$\bullet$       \\
 \textbf{D$^3$I}    &       \cite{peng2019d3i} &      &                 &       $\bullet$     &   $\bullet$  &$\bullet$ &$\bullet$     \\
\textbf{MR-GNN}     &    \cite{xu2019mr}     &     &     $\bullet$      &                &       $\bullet$      &&   $\bullet$  \\
      \textbf{SkipGNN} &     \cite{huang2020skipgnn}     &    $\bullet$       &                &    &&  &    $\bullet$   \\
\textbf{CASTER}     &    \cite{huang2019caster}      &        $\bullet$   &          $\bullet$        &        & &&     $\bullet$   \\
\textbf{DeepDrug}     &    \cite{cao2020deepdrug}    &        $\bullet$   &          $\bullet$        &        &    &&   \\
\textbf{GoGNN}     &    \cite{gognn}         &       &          $\bullet$        &  $\bullet$       &  &&   \\
\textbf{DPDDI}     &    \cite{feng2020dpddi}      &      $\bullet$     &        $\bullet$    &      $\bullet$     &    $\bullet$    &$\bullet$ & $\bullet$   \\

 \textbf{KGNN}     &    \cite{kgnn}     &      $\bullet$     &              $\bullet$    &          &     &$\bullet$ &   $\bullet$    \\
  \textbf{BiGNN}     &    \cite{bai2020bi}    &        &              $\bullet$    &     $\bullet$      &       && $\bullet$    \\
  \textbf{MIRACLE}   &   \cite{wang2021multi} &  $\bullet$         &              $\bullet$    &         &      &&  $\bullet$    \\
  \textbf{EPGCN-DS} &\cite{sun2020structure}     &                 & $\bullet$          &      && &    $\bullet$   \\
 \textbf{SumGNN}     &    \cite{yu2021sumgnn}    &      $\bullet$     &              $\bullet$    &        $\bullet$    &   &&  $\bullet$   \\
 
  \textbf{DANN-DDI}     &    \cite{dannddi}     &     $\bullet$      &          $\bullet$    &   $\bullet$    &    $\bullet$    &$\bullet$ & $\bullet$  \\
  \textbf{SSI-DDI}     &    \cite{nyamabossi}      &     $\bullet$      &          $\bullet$    &      &    &$\bullet$ &  \\
   \textbf{MTDDI}     &    \cite{mtddi2021}    &      $\bullet$  &         $\bullet$  &      $\bullet$&  $\bullet$ &$\bullet$ &$\bullet$    \\
   \textbf{MUFFIN}     &    \cite{chen2021muffin}     &       &          $\bullet$ &$\bullet$ &  $\bullet$ &$\bullet$ & $\bullet$    \\
   
    \textbf{DDIAAE}     &    \cite{dai2021drug}     &   $\bullet$     &          $\bullet$ &$\bullet$ & &&    \\
    \textbf{RWGCN}     &    \cite{feeney2021relation}    &   $\bullet$     &          $\bullet$ &&  &&     \\
        \textbf{SmileGNN}     &    \cite{han2021smilegnn}   &     &    $\bullet$        && &$\bullet$ &  $\bullet$     \\
        
        \textbf{GCN-BMP} &\cite{CHEN202047gcnbmp}&   $\bullet$     & $\bullet$   &&     && $\bullet$  \\
        \hline
         \textbf{DeepSynergy} &   \cite{preuer2018deepsynergy}      &     $\bullet$   &     $\bullet$           &    $\bullet$     &     &$\bullet$ &      \\
         \textbf{MCDC} &   \cite{mcdc}           &       $\bullet$   &     $\bullet$           &      &      & &     \\

         \textbf{DTF}&  \cite{sun2020dtf}           &       $\bullet$   &     $\bullet$           &    $\bullet$     &    &$\bullet$ &       \\
    \textbf{DeepSignalFlow}     &    \cite{zhang2021mining}    &        &       &&   &$\bullet$ &    \\
    \textbf{DeepDDS}     &    \cite{wang2021deepdds}      & $\bullet$      &  $\bullet$        &$\bullet$   &    &$\bullet$ &   \\
    \textbf{GraphSynergy}     &    \cite{yang2021graphsynergy}      & $\bullet$      &  $\bullet$        &  &   $\bullet$ &$\bullet$ & $\bullet$    \\
    \textbf{TranSynergy}     &    \cite{liu2021transynergy}    &     $\bullet$      &  $\bullet$        &  &   &&     \\
    \textbf{MatchMaker}&    \cite{brahim2021matchmaker}    & $\bullet$      &  $\bullet$        &  &   &&     \\
    \textbf{AuDNNSynergy}&    \cite{zhang2021synergistic}      &  $\bullet$      &   $\bullet$         &   $\bullet$  &  &$\bullet$ &      \\
    \textbf{AID}&    \cite{kim2021anticancer}    &  $\bullet$      &           &     &    &&    \\
    \textbf{MOOMIN}     &    \cite{rozemberczki2021moomin}      &   $\bullet$     &     $\bullet$     &&  && $\bullet$     \\
      \hline
\end{tabular}
}
\end{table}

%% file: figures/split.tex
\begin{figure}[h!]
\centering 
\includegraphics[scale=0.15]{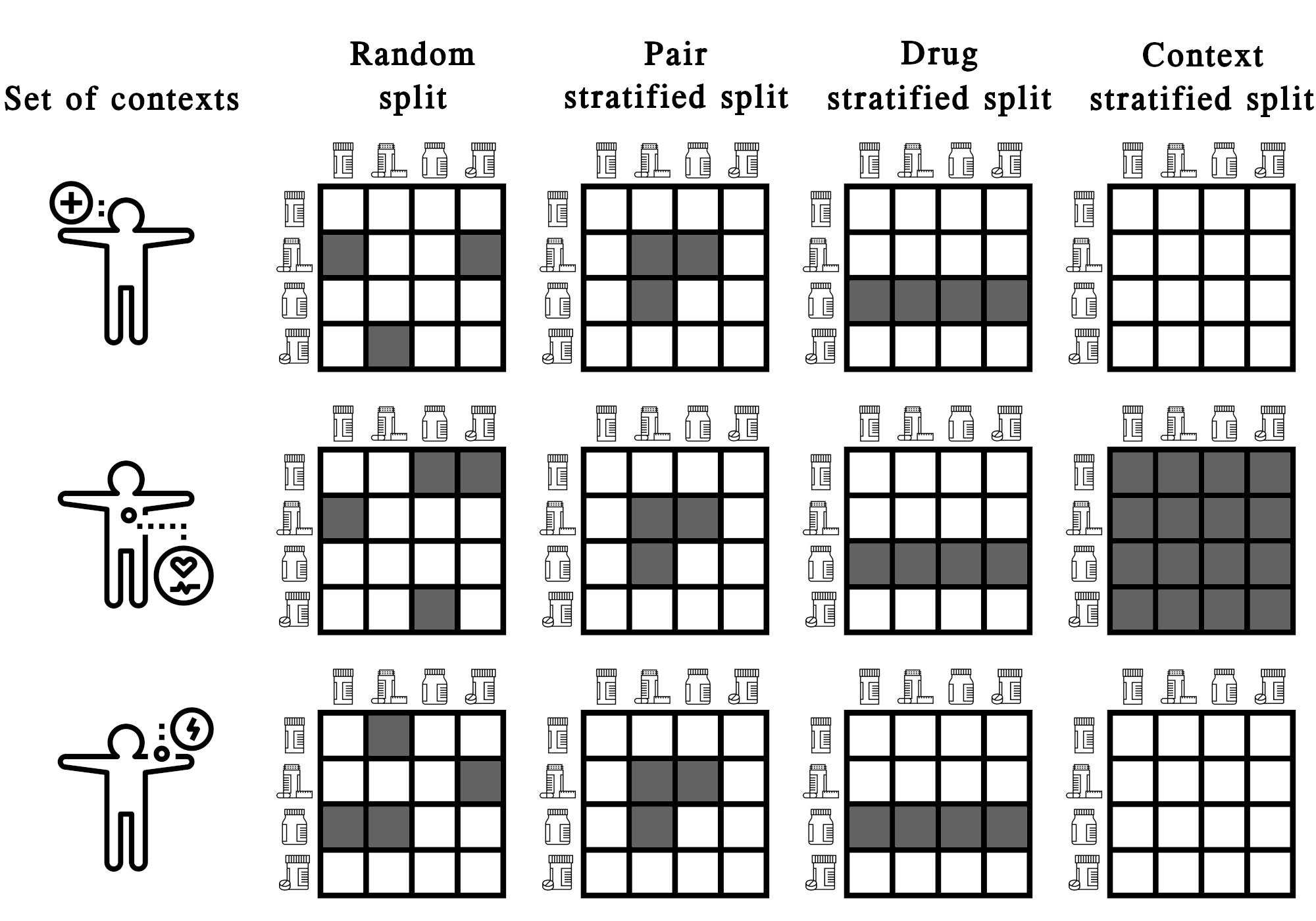}
\caption{The train and test split of drug pair scoring datasets allows for various stratified splits. Stratification of the pairs can happen across the evaluated drugs or the outcomes being tested. Drug pairs used for training the pair scoring model are denoted with white, pairs used for testing are denoted with gray.}\label{fig:splits}
\end{figure}

%% file: tables/datasets.tex
\begin{table*}[ht!]
\centering
\caption{Public drug pair scoring datasets ordered chronologically with the subtask, number of compounds, count of tested compound pairs, cardinality of the context set, licence and if commercial use is explicitly restricted.}\label{table:datasets}

{\tiny
\begin{tabular}{lccccccc}
\toprule
 \textbf{Dataset}& \textbf{Reference} & \textbf{Subtask} & \textbf{Compounds} & \textbf{Pairs}  & \textbf{Contexts} & \textbf{Licence} & \textbf{Restricted}  \\  
 \hline
 TWOSIDES & \cite{tatonetti2012data} & Polypharmacy & 1,918 & 211,990 & 12,726 & Not Specified \\
 DrugBank 5& \cite{wishart2018drugbank} & Polypharmacy & 14,575 & 365,000+ & 86+ & CC BY-NC 4.0 & $\bullet$ \\
 DeepDDI & \cite{ryu2018deep} & Polypharmacy & 1,710 & 191,995 & 86 & Not Specified \\
 TDC (TWOSIDES)& \cite{huang2021therapeutics} & Polypharmacy & 645 &  63,473 & 1,317 & CC BY 4.0\\
 TDC (DrugBank)& \cite{huang2021therapeutics} & Polypharmacy & 1,706 & 191,519 & 86 & CC BY-NC 4.0 & $\bullet$ \\
 \hline
 STITCH-CCI 5& \cite{szklarczyk2016stitch} & Interaction & 389,393 & 17,705,799 & 4 & CC BY-NC-SA 4.0 & $\bullet$ \\
 ZhangDDI & \cite{zhang2017predicting} & Interaction & 548 & 48,548 & - & Not Specified \\
 ChCh-Miner & \cite{biosnapnets} & Interaction & 1,514 & 48,514 & - & Not Specified \\
 \hline
 DCDB &  \cite{liu2010dcdb}& Synergy & 485 & 499 & -- & Not Specified \\
 DCDB 2.0& \cite{liu2014dcdb}  & Synergy &   904& 1,363 &  --& Not Specified\\
 ASDCD & \cite{chen2014asdcd} & Synergy & 105 & 215 & -- & Not Specified\\
 O'Neil & \cite{o2016unbiased} & Synergy & 38 & 583 & 39 & Not Specified\\
 NCI-ALMANAC & \cite{holbeck2017national} & Synergy & 104 & >5,000 & 60 & CC BY 4.0\\
 DrugComb &  \cite{zagidullin2019drugcomb}& Synergy &  2,276&  437,932&   93 &  CC BY-NC-SA 4.0\\
 SynergyXDB & \cite{seo2020synergxdb} & Synergy &  1,977& 22,507 & 151 & Not Specified  \\
 DrugCombDB & \cite{liu2020drugcombdb} & Synergy & 2,887 &  448,555 & 124 & Not Specified & $\bullet$ \\
 TDC (OncoPolyPharm)& \cite{huang2021therapeutics} & Synergy & 38 & 583 & 39 & CC BY 4.0\\
 DrugComb 2.0 &  \cite{zheng2021drugcomb}&  Synergy  &8,397  &  751,498& 2,320 & CC BY-NC-SA 4.0 \\
 TDC (DrugComb)& \cite{huang2021therapeutics} & Synergy & 129 & 5,628 & 29 &  CC BY-NC-SA 4.0\\
 \bottomrule
\end{tabular}
}
\end{table*}

%% file: sections/applications.tex
\section{Applications}\label{sec:applications}

In this section we introduce three key, yet currently largely, unexplored applications for the methods detailed in this review. 

\subsection{Combination Therapy for COVID-19}

One topical application of these methods is in relation to COVID-19 pandemic. Patients affected by polypharmacy of certain drug types (anti-psychotics and opiates being prominent examples) had a significantly higher chance of a negative clinical outcome from COVID-19~\cite{iloanusi2021polypharmacy,jin2021deep}. Using methods covered in this review to predict which  combinations may have a negative effect for COVID-19 patients, could enable high risk groups to seek alternative treatments, reducing the risk of a negative outcome.

\subsection{Antibiotic Evolutionary Pressure}

The prevalent use of antibiotics has resulted in microbes evolving resistance to the drugs, reducing efficacy and potentially eliminating cost effective ways of treating severe bacterial-related diseases such as Tuberculosis. Interestingly, it has been shown that the combination of different antibiotics can slow, and even reverse, this evolutionary resistance~\cite{singh2017suppressive}. However discovering these suppressive interactions using traditional methods is a complex and slow process, yet one currently unexplored using the methods covered in this review. 

\subsection{Reducing Toxicity}

Although drug combinations can result in an increase of unwanted side effects, one promising application is that the combination of two or more drugs can actually lead to a reduced level of toxicity for patients. This is due to the fact that synergistic drugs, which together posses a higher level of efficacy at targeting a certain condition, means that the levels of each individual compound can actually be lowered, reducing toxicity issues associated with higher doses~\cite{ianevski2020syntoxprofiler}. Thus, accurate prediction of synergistic drug combinations can reduce the impact of toxicity resulting from the individual compounds.


%% file: sections/discussion_and_future_directions.tex
\section{Discussion and Future Directions}\label{sec:discussion}
The body of work regarding relational machine learning for drug pair scoring primarily focuses on the design of novel architectures and applications. Our unification survey identified a number of potential shortcomings of existing approaches and venues for novel research in the domain. 
\subsection{Encoding Molecular Geometry} Our summary on the design of relational machine learning architectures for drug pair scoring tasks in Table \ref{table:general} highlighted that molecular geometry and spatial structure of the molecules is rarely encoded by existing models. Recent advances in geometric deep learning applied to chemistry \cite{qi2017pointnet,xie2018crystal,fey2018splinecnn} would allow the inclusion of geometric information which could lead to better predictive performance on the pair scoring tasks. The modularity of existing architectures makes replacing the molecular encoders with state-of-the-art geometric encoder layers a possibility.

\subsection{Higher Order Drug Combinations}
Existing research about the interactions, unwanted side effects and synergy of drugs is primarily focused on the evaluation of binary pair combinations. This is driven by the lack of datasets focused on the outcomes of using higher order drug combinations and the lack of architectures designed specifically for these higher order combinations. By using set based representation aggregation layers \cite{set2set,baek2020accurate}, the existing pair scoring models could be adapted to generate drug subset representations.

\subsection{Transfer Learning}
Self-supervised and unsupervised learning for pretraining molecular encoders is already widely used for single molecule machine learning tasks \cite{dewulf2021cold}. This provides an opportunity for pretraining the molecular encoders on single molecule tasks and fine-tuning them on the data scarce, pair scoring tasks. Another opportunity for transfer learning comes from the fact that certain pair scoring tasks have a greater quantity of labeled data available. The summary of drug pair scoring datasets in Table \ref{table:metrics} demonstrated that the \textit{drug-drug interaction prediction} task has datasets such as STITCH-CCI-5 which covers a large number of pair combinations, while the polypharmacy side effect and synergy prediction tasks have smaller databases. Pretraining models by performing drug-drug interaction prediction and fine-tuning these models for other tasks seems to be an important future research direction for training accurate, and therefore useful, models.

\subsection{Multimodal Learning}
A heterogeneous graph based representation of drugs allows for the fusion of multiple data modalities. Our survey of existing models in Table \ref{table:general} has demonstrated that only a handful of existing architectures integrates multimodal data effectively \cite{rozemberczki2021moomin,liu2021transynergy} without losing induction. Integrating multi-omics data such as proteomics, molecular structure and biological pathway information could be an important venue for designing novel pair scoring architectures.

\subsection{Software for Drug Pair Scoring}
Currently there is no dedicated open-source machine learning library which was specifically designed for solving the drug pair scoring task.  Developing a dedicated relational machine learning framework on top of existing geometric deep learning 
\cite{feylenssen2019,wang2019deep,rozemberczki2021pytorch} and deep chemistry frameworks \cite{ramsundar2019deep,korshunova2021openchem} could be an important contribution to the domain. This would require curated datasets and the architectural design of encoder, combiner, scoring layers and drug pair iterators.

%% file: sections/conclusion.tex
\section{Conclusion}\label{sec:conclusions}

We have provided an exhaustive overview of relational machine learning models designed to solve drug pair scoring tasks. We outlined a general theoretical framework which unifies the drug-drug interaction, polypharmacy side effect and drug synergy prediction tasks and created a taxonomy of models which address these. By surveying the literature, considering the architecture and evaluation of existing models, we identified key real world application areas and important directions for future research.